\documentclass[conference]{IEEEtran}
\usepackage{amsmath,amssymb,amsfonts}
\usepackage{graphicx}
\usepackage{textcomp}
\usepackage{xcolor}
\usepackage[numbers]{natbib}

\usepackage{color}
\usepackage{threeparttable}
\usepackage{algorithm}  
\usepackage{algpseudocode}  
\usepackage[capitalise]{cleveref}
\usepackage{booktabs}
\usepackage{url}

\def\BibTeX{{\rm B\kern-.05em{\sc i\kern-.025em b}\kern-.08em
    T\kern-.1667em\lower.7ex\hbox{E}\kern-.125emX}}
\begin{document}

\title{Keyphrase Extraction Using Neighborhood Knowledge Based on Word Embeddings

}

\author{Yuchen Liang\\Department of CS, RPI\\\texttt{liangy7@rpi.edu} \and
Mohammed J. Zaki\\Department of CS, RPI\\\texttt{zaki@cs.rpi.edu}}

\maketitle

\begin{abstract}
Keyphrase extraction is the task of finding several interesting phrases in a text document, which provide a list of the main topics within the document. Most existing graph-based models use co-occurrence links as cohesion indicators to model the relationship of syntactic elements. However, a word may have different forms of expression within the document, and may have several synonyms as well. Simply using co-occurrence information cannot capture this information. In this paper, we enhance the graph-based ranking model by leveraging word embeddings as background knowledge to add semantic information to the inter-word graph. Our approach is evaluated on established benchmark datasets and empirical results show that the word embedding neighborhood information improves the model performance. 
\end{abstract}

\begin{IEEEkeywords}
keyphrase extraction, graph-based model, unsupervised learning
\end{IEEEkeywords}

\section{Introduction}
\label{Introduction}

With the rapid growth of document collections available nowadays, an important task is to find a good summary of a document to facilitate knowledge acquisition. The task of keyphrase extraction involves finding informative phrases that are closely related to the main topics within the document \cite{turney2000learning,tomokiyo2003language,ding2011keyphrase,zhao2011topical}. High quality keyphrases extracted from the documents can be regarded as a concise summary of the document, which can be used for fast browsing and searching. Keyphrases can also be used to help other natural language processing (NLP) and information retrieval (IR) tasks, such as document indexing \cite{gutwin1999improving}, document clustering \cite{hammouda2005corephrase}, text categorization \cite{hulth2006study}, opinion mining \cite{berend2011opinion} and so on. 

There are generally two approaches of dealing with this problem: supervised and unsupervised \cite{hasan2010conundrums,hasan2014automatic}. In supervised learning, the task can be regarded as a binary classification problem \cite{frank1999domain,witten2005kea}. It involves finding good statistical and linguistic features and then training a binary classification model to predict whether or not the candidate phrases are keyphrases. However, supervised learning methods require human-labeled data for training, and can be expensive and inefficient when compared to the rapidly growing document collections. Besides, the model trained on a corpus from one area may not adapt well when it is applied to other text domains.

In unsupervised learning, the graph-based ranking techniques are essentially the state-of-the-art \cite{hasan2014automatic}. The basic idea of graph-based ranking model is ``voting" or ``recommendation" \cite{mihalcea2004textrank}. The general procedure is as follow: First, construct an inter-word graph based on the information within the document. Then, run some graph-based ranking model such as PageRank \cite{brin1998anatomy} or HITS \cite{kleinberg1999authoritative} to compute the score for each word, as done by many previous works on keyphrase extraction \cite{mihalcea2004textrank,wan2008single,liu2010automatic,florescu2017positionrank,litvak2008graph} . Finally, the keyphrases are selected based on the computed scores for words and phrases. A variety of graph-based ranking models have been proposed which utilize additional sources of information found to be useful for keyphrase extraction. For example, \cite{wan2008single} use neighborhood documents to strengthen the relation between two words if they co-occur in several similar documents simultaneously. \cite{liu2010automatic} decompose the document into separate topics, then compute the word score with respect to each topic and finally aggregate the score with the topic distribution of the document. \cite{florescu2017positionrank} propose a biased PageRank model by leveraging the position information of words. However, only statistical information is considered in these models. When computing the term relatedness, they use co-occurrence as the cohesion indicator of the terms. The shortcoming is that the semantic links between the words are neglected.

Statistical and semantic information can both help improve keyphrase detection models, but the key issue is how to combine them. Semantic information can help identify closely related items, which can be used to refine the weighted graph. For example, a concept may appear several times in the document but in different expressions, even though the underlying words may be closely related semantically. If we just use statistical information such as co-occurrence, such keyphrases may not be identified. Therefore,
some related works try to capture the semantic information. For example, \cite{liu2009clustering} measure the term relatedness based on the vector of Wikipedia concepts, and propose a clustering model based on that term relatedness. \cite{grineva2009extracting} exploit information extracted from Wikipedia to weight terms to determine semantic relatedness between them, which is then used for community detection for key term extraction. \cite{wang2014corpus} propose a way to compute weights in the graph-based model by leveraging word embedding vectors as the semantic information. 
%Different from their work, we propose another way of leveraging word embedding, where word embedding is %used to add in the neighborhood information. 

In this paper, we propose a way to improve the performance of graph-based ranking model by using word embeddings as the background semantic information. 
%Precisely, we use the pre-trained Word2Vec model on 100 billion words from Google News. 
Our model represents a document as a weighted graph, where vertices are the phrases and the edges measure the strength of the connection between them. We first count the co-occurrence of the phrases within a certain window size. Then, we compute the edge weights between the vertices with the help of word embeddings. 
Unlike previous approaches, we leverage the neighborhood information from word embeddings, which we find is very effective in capturing latent semantic relatedness.
The underlying intuition here is that: suppose in the graph, the phrase A co-occurs with word B, C and D. If B and C are semantically related, the connection between (A, B) and (A, C) should be stronger. We construct the weighted graph by leveraging this kind of semantic relatedness. Then, we adopt a Biased PageRank method to rank the phrases based on their PageRank score. Finally, we propose an ensemble approach to get the output keyphrases list. Our model is evaluated on several benchmark datasets, and the evaluation results show improved performance when using the word neighborhood information.
%better precision and recall values compared to the baseline algorithms. 

\section{Related Work}
\label{related work}

Generally, the keyphrase extraction problem can treated as either a supervised or unsupervised task.  

Supervised approaches usually involve finding good features from the text document and then training a binary classification model \cite{frank1999domain,witten2005kea}. Different features and classification models have been used. The KEA system \cite{witten2005kea} uses Tf-idf and first occurrence attribute as features during the training and extraction process, and uses Naive Bayes as the classifier. \cite{hulth2003improved} proposes a model which includes some linguistic knowledge. Four features are used in the model including the within document frequency, the collection frequency, the position of the first occurrence and POS tags, and bagging is used to train the classifier. Other classification methods that have been used include decision trees \cite{turney2000learning}, maximum entropy \cite{yih2006finding,kim2009re}, and support vector machines \cite{jiang2009ranking,lopez2010humb}. Recently, people begin to use deep neural network to extract the keyphrases to avoid the burden of hand-craft features. \cite{meng2017deep} propose an end-to-end keyphrases generation model which leverages the encoder-decoder framework, copy mechanism is used in the model to balance the keyphrases which are extracted or generated.  \cite{alzaidy2019bi} treat the keyphrases extraction as sequence labeling problem, propose a model combining label dependencies learned by conditional random fields with the semantic meaning among the neighbors learned by  bidirectional long short term memory networks.

For unsupervised approach, Tf-idf is a simple, though still effective baseline. However, graph-based ranking models are considered to be the state-of-the-art \cite{hasan2014automatic}. The TextRank method \cite{mihalcea2004textrank} leverages PageRank in the keyphrase extraction domain, where a graph is first built on the words within the document. After that work, a variety of graph-based ranking models which focus on adding different types of information have been proposed. \cite{wan2008single} propose SingleRank, where the edge weight is set to be the number of co-occurrences between the words, and in the keyphrase selection step they use the sum of scores of the words contained in a phrase. ExpandRank \cite{wan2008single} uses neighborhood documents to strengthen the connection between the words that co-occur in similar documents. \cite{gollapalli2014extracting} integrate information from citation networks. \cite{wang2014corpus} propose a graph-based ranking model leveraging the embeddings, with the weight between the words coming from the product of two scores: (a) the attraction force, which uses the words frequency as well as the distance between the words in the embedding space, and (b) the dice coefficient \cite{dice1945measures,stubbs2002two} to compute the similarity. \cite{florescu2017positionrank} take into account the position information of the words, and use a biased PageRank model to accommodate the position information, where the words appearing earlier and frequently tend to get a larger weight. \cite{mahata2018key2vec} propose a graph-based ranking model which uses the phrase embedding to capture the semantic relatedness between the phrases, the weights of the graph consist of cosine similarity between the phrases as well as the co-occurrence score.
 
Several other approaches try to first identify the topics within the document, and then extract the keyphrases based on the topic to make sure the extracted keyphrases get a good coverage of all the topics for each document. Existing approaches include clustering \cite{liu2009clustering,bougouin2013topicrank}, where they use different methods to group the words based on various distance metrics, and then extract representative phrases from each group as the keyphrases. \cite{liu2010automatic} combine topic modeling with PageRank. Specifically, they first decompose the document into a combination of several topics. Each topic has its own word distribution, which is used to compute a biased PageRank for each topic. The final word score is computed by the aggregation of the scores in all the topics.

RAKE \cite{rose2010automatic} is another unsupervised approach for keyphrase extraction. The idea is to cut the document into several chunks with the help of a stop word list as well as delimiters. Those chunks are regarded as the candidate phrases. The word score is then computed based on the candidate phrases, where several score functions are proposed such as the frequency of the word, the degree of the word and so on. Finally, the candidate phrase score is the sum of the score for each word.

\section{Proposed Approach}
\subsection{Overview}
In the graph-based ranking model, usually we use co-occurrence  within a certain window size as a cohesion indicator between words/phrases. The more times two words/phrases co-occur, the stronger the connection between these two words, the more likely are they to receive higher PageRank scores. However, a concept can be expressed in different forms, and using co-occurrences without semantics can lead to information loss. For example, suppose phrase A co-occurs with phrase B, C, D one time each, and we have the background knowledge that the semantic meaning of B is very close to C. Running PageRank on the co-occurrence graph will yield the same scores for B, C and D. However, intuitively B and C should receive a higher score and the connection between A and the concept represented by B and C should be more strong. Our approach is designed to address this problem. For the weight of a given edge $AB$, we consider not only the co-occurrence between A and B, but also the co-occurrence between A's neighborhood set and B, as well as the co-occurrence between B's neighborhood set and A. With the help of this neighborhood set, we can strengthen the inter-phrase relationship, which cannot be captured using statistical information alone. 
%ExpandRank \cite{wan2008single} has a similar idea, where they use the neighborhood %document word relationships as the supplement information to strengthen the words %connection. 

\begin{algorithm}[!htb]  
	\caption{Keyphrase extraction based on neighborhood information}  
	\label{alg:Framwork}  
	\begin{algorithmic}[1]  
		\Require  
		A text document
		\Ensure  
		Extracted keyphrases for the document
		\State Do tokenization %(which cut the document into words and punctuations) 
		and POS tagging for the document. Candidate phrases with certain POS tags are added to the vertex set
		\State \textbf{Neighborhood Construction:} For each phrase $v_0$, create a neighborhood set $D_0 = \{v_0^1, v_0^2, \cdots v_0^k\}$ of size $k$, which contains the phrases that are most similar to $v_0$ using word embeddings
		\State Construct a Weighted Undirected Graph with the edge weight between vertex $v_i$ and vertex $v_j$ based on the co-occurrence between $v_i$ and $v_j$, and also the co-occurrence between $v_i$ and other phrases in $v_j$'s neighborhood set $D_j$, as well as the co-occurrence between $v_j$ and other phrases in $v_i$'s neighborhood set $D_i$
		\State Run Biased PageRank algorithm to get the score for each vertex  
		\State Use ensemble approach to aggregate the final output   
	\end{algorithmic}  
\end{algorithm} 

A high-level outline of our keyphrase extraction approach is 
shown in Algorithm~\ref{alg:Framwork}. Details of the various steps are given below.

\subsection{Candidate phrase}

We first extract the candidate phrases as the vertices for the weighted graph, which are extracted from the document according to the following pattern: $(adjective)*(noun)+$, which means zero or more adjectives followed by at least one noun.

\subsection{Word Embeddings}

Word embeddings project the words into a vector space so as to facilitate the extraction of latent similarity or semantics. Models such as the bag-of-words tend to give each word a one hot representation, which suffers from the curse of dimensionality, since the vocabulary can be very large. On the other hand,
Word2Vec \cite{mikolov2013linguistic} is a neural-network based model that uses the neighborhood context to capture the word distribution, and has been shown to be effective in capturing the latent semantic meaning of the words.
In this work, we use the pre-trained Word2Vec model on 100 billion words from Google News as the background information to calculate the word similarity (the embedding dimensionality is 300). From the word embedding to the phrase embedding, for a certain phrase, we sum up all the word embedding scores within the given phrase as the phrase embedding value.
%For example, \cite{mikolov2013linguistic} show some of vector embedding results: $vec(king) %- vec(man) + vec(woman) \approx vec(queen)$, and $vec(apple) - vec(apples) \approx vec(car) %- vec(cars)$.

\subsection{Neighborhood Construction}

For a given phrase $v_0$, we find the $k$ nearest neighbors which are semantically related to this phrase. The similarity between $v_i$ and $v_j$ is calculated using the cosine similarity of these two phrases in their embedding space:

\begin{equation*}
%\label{E1}
sim(v_i, v_j) = \frac{\vec{v}_i\cdot \vec{v}_j}{\|\vec{v}_i\| \times \|\vec{v}_j\|}
\end{equation*} 

We construct the neighborhood set of $v_0$ based on threshold, cosine similarity exceeds certain threshold will be added into $v_0$ neighborhood set. Cosine similarity also serves as a confidence value for the semantic connection between the phrases.

\subsection{Graph-based Ranking}
\subsubsection{Graph Construction}
For a given document, we use a undirected graph $G = (V, E)$ to model the relationship between the phrases. We choose each vertex to be a candidate phrase, which is restricted to certain pattern. We use the number of co-occurrences within a specific window size $w$ to indicate the cohesion relationship between phrases. For edge $e_{ij}$ between phrases $v_i$ and $v_j$, we consider not only the co-occurrence between the phrase $v_i$ and $v_j$, but also the co-occurrence between the phrase $v_i$ and the phrases in the neighborhood set of $v_j$, as well as the co-occurrence between the phrase $v_j$ and the phrses in the neighborhood set of $v_i$:

\begin{align*}
%\label{E2}
w(v_i, v_j) & = \sum_{v_k \in D_j}sim(v_j, v_k)\times  count(v_i, v_k)\\
 & + \sum_{v_l \in D_i}sim(v_l, v_i)\times  count(v_l, v_j)
\end{align*} 
where $count(v_i, v_k)$ is the number of co-occurrences between phrase $v_i$ and $v_k$, and $D_j$ is the neighborhood set of the phrase $v_j$

\subsubsection{Biased PageRank}

Next, we use a biased PageRank approach to compute the score for each vertex. Formally, in the graph G = (V, E) where V is the set of all vertices and E is the corresponding edge set, let $C(v_i)$ indicate the set of vertices that are connected to $v_i$. Then, the score of $v_i$ is calculated as

\begin{align*}
%\label{E3}
%\small
S(v_i) & = (1 - d)\times r_i^*\\
 & +
d \times\sum_{v_j\in C(v_i)}\frac{w(v_j, v_i)}{\sum_{v_k \in C(v_j)}w(v_j, v_k)}S(v_j)
\end{align*}
where $w(v_j, v_i)$ denotes the weight between vertex $v_j$ and $v_i$. The parameter $d$ is the damping factor, usually chosen to be 0.85. For the random jump vector $r_i^*$, we follow \cite{florescu2017positionrank}, considering the position information. Specifically, for each phrase $v_i$, we let $\tilde{r_i}$ to be the sum of the inverse position of the sentences where $v_i$ appears. For example, if $v_i$ appears at both the 3rd and 10th sentences, then $\tilde{r_i} = \frac{1}{3} + \frac{1}{10} = 0.433$. The intuition is that the phrases which appear frequently and earlier in the document tend to be more important. $\tilde{r_i}$ is then normalized. Different from PositionRank, here we also use Neighborhood information to construct the biased weighted vector. for each phrase $v_i$, we compute the final weight $r_i^*$ not only based on its own position information in the document (which is $\tilde{r_i}$), but also consider other phrases' position information in $v_i$'s neighborhood set:

\begin{equation*}
%\label{E4}
r_i^* = \sum_{v_k \in D_i}sim(v_i, v_k)\tilde{r_k}
\end{equation*} 

The intuition is that if some other similar phrases appears more frequently and earlier, it should increase the chance that the phrase itself is important.

In the experiments, the vertex scores are computed recursively until the difference between two consecutive iterations is small (e.g., less than 0.0001) or we reach the max number of iterations (e.g., 100).

\subsection{Rank aggregation}

After we compute the PageRank score for the vertices, we can sort the candidate phrases decreasingly according to their PageRank score. Instead of directly output it as the final result, here we propose an ensemble model, which aggregates our model with some other model which takes consideration of the other documents in the corpus, since our model is document based, it may lose some "global" information(for example, some high ranked phrases may also high ranked in many other documents, but they are too common to be the keyphrases). First, we propose a naive approach to combine our model with TF-IDF, the general procedure is (a) select top 25 candidate phrases for each model to construct candidate list $l_a$ and $l_b$ (b) the keyphrase should be in both methods' candidate list (c) if a single word phrase appears as parts of another multi-word phrase in the candidate list, then filter out the single word phrase (d) the relative ranking for each candidate list is unchanged. the pusedo code is as below:

\begin{algorithm}[!htb]  
	\caption{ensemble approach}  
	\label{alg:Ensemble}  
	\begin{algorithmic}[1]  
		\Require  
		Candidate list $l_a$, $l_b$
		\Ensure  
		Final output $l_{out}$
		\State $l_{out}$ = []
		\For{$i=1$ to $m$}  
            \If {$l_a[i]$ in $l_b$ \textbf{and} $l_a[i]$ not in $l_{out}$}
                \State $l_{out}.append(l_a[i])$
                \If {$l_a[i]$ is single word phrase}
                    \If {$l_a[i]$ has counterpart multi-word phrase in $l_a$ or $l_b$}
                        \State $l_{out}.pop()$
                    \EndIf
                \EndIf
            \EndIf
            \If {$l_b[i]$ in $l_a$ \textbf{and} $l_b[i]$ not in $l_{out}$}
                \State $l_{out}.append(l_b[i])$
                \If {$l_b[i]$ is single word phrase}
                    \If {$l_b[i]$ has counterpart multi-word phrase in $l_a$ or $l_b$}
                        \State $l_{out}.pop()$
                    \EndIf
                \EndIf
            \EndIf
        \EndFor
	\end{algorithmic}  
\end{algorithm} 

Besides, we also try some rank aggregation methods. We use Kemeny-Young rank aggregation method \cite{kemeny1959mathematics, young1988condorcet}, which minimizes the total kendall-tau distance for the ranking score. In our implementation, we use integer program given by \cite{conitzer2006improved} to approximate the result. Finally, after rank aggregation, we output top $m$ phrases in the combined candidate list as the keyphrases, where $m$ equals to the average golden keyphrses per document.

\section{Experimental Setup}
\subsubsection{Datasets}
To compare our model with existing approaches, we use three benchmark datasets for experiments (see Table \ref{tab:stats}).
The \textbf{Inspec} dataset \cite{hulth2003improved} is a collection of 2000 paper abstracts as well as the human annotated keyphrases. The size of each abstract is relatively small, with the tokens per document fewer than 200 on average \cite{hasan2014automatic}. 
%The 2000 abstracts are divided into 1000 for training, 500 for validation and 500 for %testing by author. Since our method is unsupervised learning model, we only need 500 test %data for experiment and model comparison. Besides, 
This dataset has two types of keyphrases: controlled keyphrases and uncontrolled keyphrases. The controlled keyphrases are selected from a pre-defined dictionary and the uncontrolled keyphrases are the freely assigned keyphrases. Following previous work, we use uncontrolled keyphrases for evaluation, and only the keyphrases appearing at least once in the document are considered.    
The \textbf{DUC2001} challenge dataset \cite{over2001introduction} contains 308 news articles, along with the human annotated keyphrases built by Wan and Xiao \cite{wan2008single}. The size of each news article is 740 words on average. Thus, both the document size and the domain are quite different from Inspec. 
The \textbf{NUS} corpus \cite{nguyen2007keyphrase} contains 211 scientific papers from various disciplines. Each paper has its own author and reader annotated keyphrases. The author-provided keyphrases are used as golden-standard for comparison, and
we use the paper abstracts to extract the keyphrases.
%Following the PositionRank, we only use the abstract to extract the keyphrases in this dataset. 
  
\begin{table*}[!htb]
\centering
	\caption{Summary of Datasets}
	\begin{tabular}{ccccc}
		\toprule
		Source & Dataset & Documents & Tokens/doc & Keyphrases/doc\\
		\midrule
		Paper abstracts & Inspec \cite{hulth2003improved} & 2000 & $<$200 & 10\\
		News articles & DUC2001 \cite{wan2008single} & 308 & $\approx900$ & 8\\
		Scientific papers & NUS corpus \cite{nguyen2007keyphrase} & 211 & $\approx8\text{K}$ & 11\\
		\bottomrule
	\end{tabular}
\label{tab:stats}
\end{table*}

\subsubsection{Baselines}

We compare our model performance with several previous approaches. The first baseline is Tf-idf, which can achieve relatively good results \cite{hasan2010conundrums}. The method chooses the candidate phrases according to their frequency in the document, as well as the inverse of their frequency in other documents. Several other graph-based ranking models are also compared including TextRank \cite{mihalcea2004textrank}, SingleRank \cite{wan2008single}, ExpandRank \cite{wan2008single} and PositionRank \cite{florescu2017positionrank}. All these methods use undirected graphs, and words with certain POS tags (adjective and noun) are added into the vertices list. SingleRank use co-occurrence of the words within certain a window size to model the edges of the graph.
%TextRank uses random weights whereas SingleRank uses the number of co-occurrences as the weight. 
ExpandRank considers not only the word co-occurrence within the document, but also the word co-occurrence in neighboring documents. 
%the product of two coefficients to be the edge weights of the graph: (a)the attraction %force, which uses the word frequency as well as the distance between the words in the %embedding space, and (b)the dice coefficient \cite{dice1945measures,stubbs2002two}, which %uses the co-occurrence frequency of the words and the frequency for each single word. 
SingleRank and ExpandRank use PageRank to compute the score for each vertex and extract the final keyphrases according to the computed word score.
On the other hand, PositionRank uses biased PageRank, which leverages the position information. In the keyphrases extraction step, 
TextRank is different from other models, they select top-k words as the keyphrases, and adjacent keyphrases are collapsed into a multi-word phrase, whereas other approaches first select the candidate phrases according to some pattern (e.g., $(adjective)*(noun)+$), and then compute the phrase score as the sum of the words score. Finally, they select top-k scoring phrases as the keyphrases.

\subsubsection{Evaluation Measures}

We evaluate our results in terms of precision, recall and f-measure under various parameter settings, comparing the extracted phrases with human annotated phrases for each dataset, after stemming.

\section{Results and Discussion}
There are four hyper-parameters in our model.
%(if you use different threshold to construct neighborhood set for weighted graph and biased %weighted vector, then there will be four hyper-parameters). 
The first  is the window size $w$, which controls the scope when we count the word co-occurrence. The threshold we use when we construct the neighborhood set of the words uses two parameters: $t_1$ denotes the threshold used to construct neighborhood set for weighted graph, and $t_2$ denotes the threshold used to construct neighborhood set for the biased weight vector. Finally, we have the number of keyphrases $m$ to output.

\begin{table*}[!ht]
	\centering
		\caption{Performance Comparison}
		\begin{tabular}{cccccc}
			\toprule
			 & Algorithm & Parameter & Precision & Recall & F-score\\
			\midrule
			Inspec & TF-IDF & w = 10, m = 8 & 37.8 & \textbf{38.7} & 38.2 \\
			       & TextRank & w = 10, m = 8 & 34.6 & 35.3 & 35.0 \\
					  & SingleRank & w = 10, m = 8 & 36.8 & 37.7 & 37.2 \\
					  & ExpandRank & w = 10, m = 8 & 36.3 & 37.3 & 36.8 \\
					  & PositionRank & w = 10, m = 8 & 36.5 & 37.4 & 36.9 \\
                      & Our Model(1) & w = 10, m = 8, $t_1$ = 0.7, $t_2$ = 0.7 & \textbf{39.7} & 37.9 & \textbf{38.7} \\
					  & Our Model(2) & w = 10, m = 8, $t_1$ = 0.7, $t_2$ = 0.7 & 38.5 & 37.0 & 37.7 \\
		
			\midrule
			DUC2001 & TF-IDF & w = 10, m = 8 & 26.5 & 26.6 & 26.5 \\
			& TextRank & w = 10, m = 8 & 23.3 & 23.4 & 23.4 \\
			& SingleRank & w = 10, m = 8 & 25.6 & 25.6 & 25.6 \\
			& ExpandRank & w = 10, m = 8 & 27.5 & \textbf{27.6} & 27.5 \\
			& PositionRank & w = 10, m = 8 & 26.5 & 26.5 & 26.5 \\
            & Our Model(1) & w = 10, m = 8, $t_1$ = 0.7, $t_2$ = 0.7 & \textbf{33.4} & 24.5 & \textbf{28.3} \\
            & Our Model(2) & w = 10, m = 8, $t_1$ = 0.7, $t_2$ = 0.7 & 27.2 & 27.2 & 27.2 \\
			
			\midrule
		    NUS corpus  & TF-IDF & w = 10, m = 3 & 9.1 & 10.7 & 9.8 \\
		    & TextRank & w = 10, m = 3 & 7.2 & 8.6 & 7.9 \\
			& SingleRank & w = 10, m = 3 & 8.9 & 10.5 & 9.6 \\
			& ExpandRank & w = 10, m = 3 & 10.0 & 11.8 & 10.8 \\
			& PositionRank & w = 10, m = 3 & 11.2 & 13.3 & 12.2 \\
            & Our Model(1) & w = 10, m = 3, $t_1$ = 0.7, $t_2$ = 0.7 & \textbf{15.2} & \textbf{18.0} & \textbf{16.5} \\
            & Our Model(2) & w = 10, m = 3, $t_1$ = 0.7, $t_2$ = 0.7 & 14.9 & 17.6 & 16.1 \\
		
			\bottomrule
		\end{tabular}
		\begin{tablenotes}
			\item \textbf{m}: number of top ranked phrases,  \textbf{w}: window size, \textbf{$t_1$}: neighborhood threshold used to construct the weighted graph (\textbf{NA} means do not use neighborhood information here, \textbf{$t_2$}: neighborhood threshold used to construct the biased weighted vector (\textbf{NA} means do not use biased PageRank here)
		\end{tablenotes}
		\label{tab:baseline}
\end{table*}

\subsection{Comparative Performance}
We compare our model with previous approaches. 
Table \ref{tab:baseline} shows the precision, recall and F-score values of our model on the Inspec, DUC2001 and Nguyen datasets.
For fairness, we obtained the implementation for PositionRank as well as several other baselines from the authors of PositionRank \cite{florescu2017positionrank} to do the replication study. All models use the same experimental setup, that is, we use the same values of the hyperparameters for all methods ($w$ and $m$) and the same approach to keyphrase formation(here we also applied the single-word phrase filtering strategy to all the baseline for fairness consideration). No attempt has been made to optimize our results by hyperparameter search. Here, we choose $m$ to be the average number of golden keyphrases per document for each dataset.

OurModel(1) indicates the neighborhood information model combined with Tf-idf using our own designed combining approach. OurModel(2) indicates the Kemeny-Young rank aggregation of the neighborhood information model, Tf-idf and the PositionRank. Based on the result, we can see our model consistently outperforms or yields competitive results compared to previous methods across the three datasets. Besides, we can find when several different base methods which capture different aspects of the data are aggregated together, the ensemble approach is more powerful. Also, one advantage is that our method doesn't heavily rely on the external resources, except for pre-trained Word2Vec embeddings. 

\subsection{Conclusions}
This paper proposes a novel approach for keyphrases extraction which leverages word embeddings to add latent semantic links between the words, so as to be able to create a more accurate weighted graph for a document. The paper also proposes an ensemble approach to aggregate some existing models, which can help us achieve better results. In future work, we plan to refine our neighborhood model and conducting more thorough replication studies on a wider variety of datasets.

\bibliographystyle{IEEEtranN}
\bibliography{keyphrase}

%\appendix
\end{document}